\documentclass[letterpaper, 10 pt, conference]{ieeeconf}    

\IEEEoverridecommandlockouts                                

\usepackage{hyperref}
\usepackage{url}
\usepackage{graphics} 
\usepackage{times} 
\usepackage{booktabs}       
\usepackage{amsfonts}       
\usepackage{nicefrac}       
\usepackage{microtype}      
\usepackage{algorithm,multirow,xcolor}
\usepackage{algorithmicx}
\usepackage{algpseudocode}
\usepackage{mathtools}
\usepackage{graphicx}
\usepackage{cases}
\usepackage{array}
\usepackage{color}
\usepackage{float}
\usepackage{amssymb}
\usepackage{cite}
\usepackage{bm}

\usepackage{wrapfig}
\usepackage{subfigure}
\usepackage{amsmath}
\usepackage{amsthm, amssymb}
\theoremstyle{definition}

\usepackage{soul}

\usepackage{makecell}
\usepackage{graphicx} 
\usepackage{subfigure}
\usepackage{epstopdf}
\usepackage{cite}
\usepackage{enumerate}
\usepackage[skip=5pt]{caption}

\title{\LARGE \bf
HAC-LOCO: Learning Hierarchical Active Compliance Control for Quadruped Locomotion under Continuous External Disturbances
} 

\author{
Xiang Zhou$^*$, Xinyu Zhang$^*$, Qingrui Zhang
  \thanks{$^*$These authors contributed equally to this work. The authors are with the School of Aeronautics and Astronautics, Shenzhen campus of Sun Yat-sen University, Shenzhen 518107, P.R. China. Correspondence to Qingrui Zhang (zhangqr9@mail.sysu.edu.cn)
  Videos available at https://b23.tv/KoX6vzm}%
  }

\begin{document}

\maketitle
\thispagestyle{empty}
\pagestyle{empty}


\begin{abstract}
Despite recent remarkable achievements in quadruped control, it remains challenging to ensure robust and compliant locomotion in the presence of unforeseen external disturbances. Existing methods prioritize locomotion robustness over compliance, often leading to stiff, high-frequency motions, and energy inefficiency. This paper, therefore, presents a two-stage hierarchical learning framework that can learn to take active reactions to external force disturbances based on force estimation. In the first stage, a velocity-tracking policy is trained alongside an auto-encoder to distill historical proprioceptive features. A neural network-based estimator is learned through supervised learning, which estimates body velocity and external forces based on proprioceptive measurements. In the second stage, a compliance action module, inspired by impedance control,  is learned based on the pre-trained encoder and policy. This module is employed to actively adjust velocity commands in response to external forces based on real-time force estimates. With the compliance action module, a quadruped robot can robustly handle minor disturbances while appropriately yielding to significant forces, thus striking a balance between robustness and compliance. Simulations and real-world experiments have demonstrated that our method has superior performance in terms of robustness, energy efficiency, and safety. Experiment comparison shows that our method outperforms the state-of-the-art RL-based locomotion controllers. Ablation studies are given to show the critical roles of the compliance action module.
\end{abstract}

\section{Introduction}

Quadruped robots are capable of adapting to a wide variety of terrains  \cite{Lee2020Learning, Choi2023Learning}, enabling agile locomotion in complex environments  \cite{Margolis2024Rapid}. Recent progress in robot hardware and locomotion control has witnessed the prevalent applications of quadruped robots in diverse industries. Despite the advancements, most existing methods focus on robust locomotion control in unstructured terrains, so they lack compliance with unexpected disturbances  \cite{Hartmann2024Deep, Lee2022Deep}. However, a safe and durable robotic system should actively adjust its posture and speed in response to external forces rather than brutally rejecting them, resulting in so-called compliance behavior.

Active compliant locomotion offers distinct advantages over its robust counterpart in terms of running safety, energy efficiency, and human-robot interaction. This preference for compliance over robustness is evident in various animals. For example, horses have evolved the capability to respond compliantly to external disturbances for survival. They possess a sophisticated system for sensing external forces and employ a layered strategy in their responses\cite{Morton_2006_Cerebellar,Merel2019Hierarchical}. When subject to minor disturbances, like a gentle kick, horses leverage spinal reflexes to automatically adjust limb stiffness, effectively absorbing the impact. However, when confronted with persistent forces, such as being guided by a human, horses engage in adjusting their speed and direction accordingly. These adaptive behaviors demonstrate a delicate balance between robustness against minor disruptions and compliance with sustained perturbations. By integrating similar compliance mechanisms into robots, we can enhance their capacity to handle external disturbances, thereby mitigating the risk of damage while improving overall performance.

\begin{figure}[!t]
 \centering
 \includegraphics[width = \linewidth]{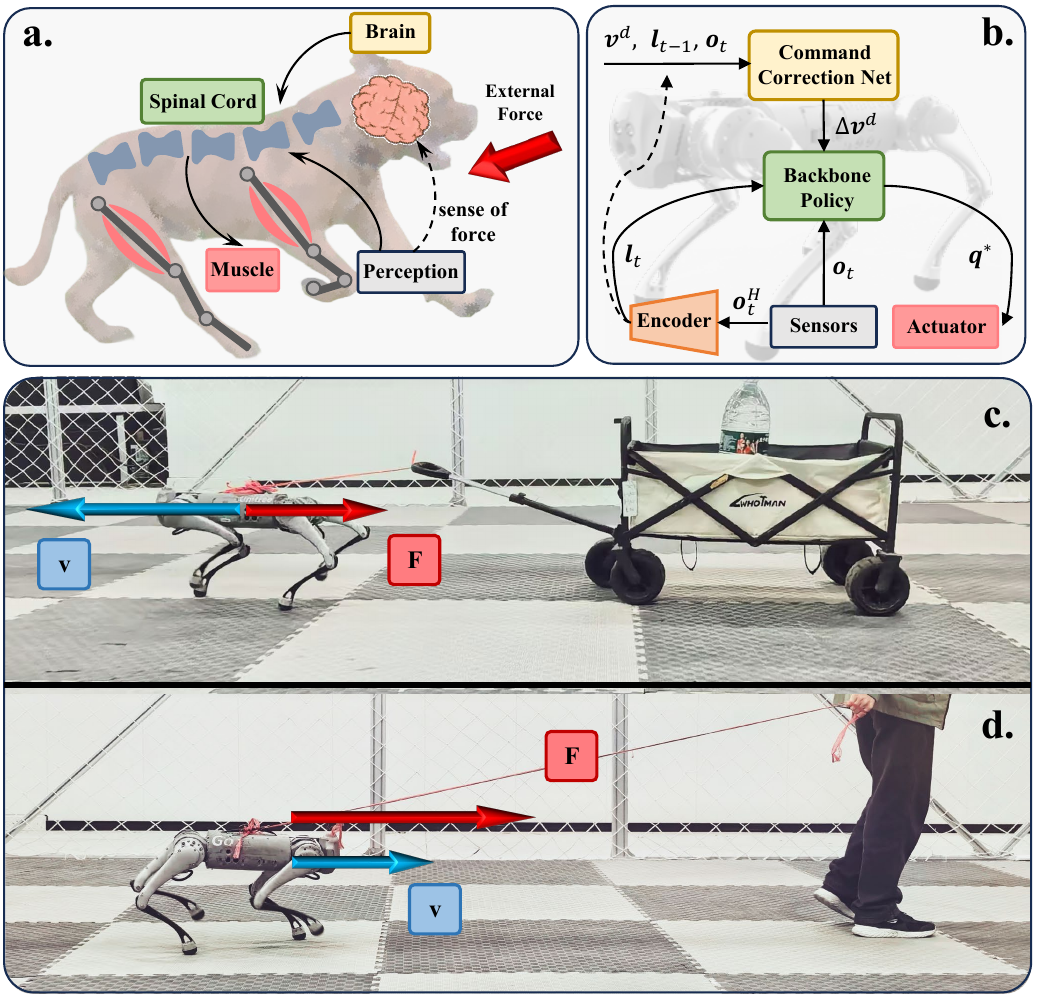}
 \caption{(a) Animals can respond with a layered strategy by sensing external forces.
(b) The proposed HAC-LOCO framework with an active compliance control method. (c) and (d) Versatile compliant behavior is generated by the HAC-LOCO algorithm.
}
 \label{fig: framework and snapshot}
\end{figure}
Despite its advantages, active compliant locomotion poses a significant challenge for quadruped robots navigating various terrains. Existing reinforcement learning (RL)-based control methods, typically tailored for optimizing task-specific rewards, tend to prioritize steady and robust locomotion\cite{Hwangbo2019Learning,Lee2020Learning,Margolis2023Walk}. However, an emphasis on robustness often results in high motion stiffness, causing the robot to exhibit high-frequency movements when faced with unexpected disturbances. These abrupt motions lead to excessive torques that surpass the motor's limits or even lead to damage. Current RL-based controllers also lack the flexibility to adapt to human intervention, raising the risk of accidents in human-robot interactions. Hence, the limitation of the state-of-the-art design underscores the need for advancements in control to strike a balance between robust locomotion and compliance behavior, ensuring safe and effective traversal of diverse terrains while maintaining adaptability to external influences.

Traditional compliance control methods, such as impedance and admittance control\cite{Zhang2024Whole}, aim to characterize the dynamic interactions between a robot and its surroundings to achieve compliance with external forces. These methods typically rely on accurate disturbance estimators to detect external forces. While effective in structured environments, they depend on precise dynamic models and involve intricate design processes, limiting their performance in dynamic and unpredictable settings. In contrast, model-free learning-based approaches, \emph{e.g.} deep neural network  \cite{Zhang2024SYNLOCOVE}, have the capability to directly estimate external disturbances based on historical sensory data, offering greater adaptability in complex environments. Hence, it is possible to emulate the performance of traditional impedance control methods by integrating a disturbance estimation module with a compliance controller through RL. This integration could pave the way for more robust and flexible control strategies that can effectively handle external disturbances in varied and dynamic environments.


In this work, we develop a hierarchical active compliance control framework called HAC-LOCO. This architecture draws inspiration from the layered response strategies of animals to achieve active compliant robot locomotion. Our framework decouples the external force estimation module from the compliance control module and follows the "sense-plan-act" paradigm. HAC-LOCO enables the robot to decide whether to resist or comply with disturbances, offering a more flexible and adaptive solution for compliant locomotion. The main contributions of this paper are threefold:

\begin{enumerate}
    \item A hierarchical reinforcement learning framework is proposed to achieve active compliance control for quadruped locomotion. This framework results in robust behavior against impact force disturbances and compliant locomotion to persistent external disturbances. Experiments validate that this framework enables the robot to exhibit natural compliance responses to disturbances, enhancing safety, stability, and energy efficiency.

    \item A learning-based estimator is designed to extract features from historical sensory feedback. This estimator merges an AutoEncoder network with explicit force and velocity estimation networks. Ablation studies demonstrate this estimator's capability to accurately estimate disturbances, allowing the robot to respond promptly to unexpected disruptions. Experimental results showcase that this proposed estimator enhances the robot's compliance, stability, and energy efficiency in complex environments.
    
    \item  Inspired by conventional impedance control, a lightweight learning-based locomotion compliance module has been developed. This module enables the robot to display compliant behaviors in reaction to persistent external disturbances. It is easy to update the compliance module to adjust the robot's impedance without worrying about retraining the whole HAC-LOCO framework. Hence, this design provides impedance-variable control, allowing for explanation and analysis.

\end{enumerate}

The remainder of this paper is organized as follows. Section \ref{sec: Related Works} provides a summary of related works on policy adaptation methods for dynamic environments and learning-based compliant control. Section \ref{sec: Methodology} introduces the proposed hierarchical reinforcement learning architecture, along with its training details. Section \ref{sec: Results} presents the experimental setup, results, and a comprehensive comparative analysis. Finally, Section \ref{sec: Conclusion} concludes the paper and outlines potential directions for future research.


\section{Related Works}\label{sec: Related Works}

\subsection{Policy adaptation for dynamic environments}
To actively adapt to dynamic environments, numerous studies focus on training policies that can sense and respond to environmental changes in real-time. One promising approach, known as privileged learning or learning by cheating  \cite{Chen2020Learning}, utilizes a teacher-student architecture to train environment-adaptive policies. In this framework, the student policy is trained under the supervision of a teacher policy that has access to privileged information. Several studies have successfully employed this paradigm to achieve improved performance. Lee \emph{et. al.} utilized privileged information to facilitate blind locomotion adaptation to challenging terrains  \cite{Lee2020Learning}. RMA leverages privileged learning to enable policies to adapt to varying motor properties and ground friction  \cite{Kumar2021RMA}. PA-LOCO incorporates a multi-encoder architecture designed to process various types of privileged information, such as trunk velocity, external disturbances, and terrain characteristics, thereby preventing the overlap of these information sources and enabling perturbation-adaptive locomotion  \cite{Xiao2024PALOCO}. DreamWaQ introduces a context-assisted estimation network based on variational autoencoders (VAEs), which guides the encoder to extract key features from historical data by predicting future states  \cite{Nahrendra2023DreamWaQ}. The Hybrid Internal Model (HIM) treats the robot’s external states as disturbances and uses a hybrid internal embedding to estimate these disturbances. This embedding is optimized through contrastive learning to align closely with the robot’s successor state, allowing for a more natural response  \cite{Long2023HIM}. Additionally, the concurrent teacher-student framework (CTS) trains both the teacher and student policies simultaneously within a reinforcement learning paradigm, demonstrating greater robustness and agility in locomotion compared to traditional two-stage privileged learning approaches  \cite{Wang2024CTS}. While these methods enable robots to adapt to dynamic environments with limited proprioceptive feedback, the existing works failed to address the challenge of adapting to persistent external disturbances.

\begin{figure*}[!t]
 \centering
 \includegraphics[width = 1\linewidth]{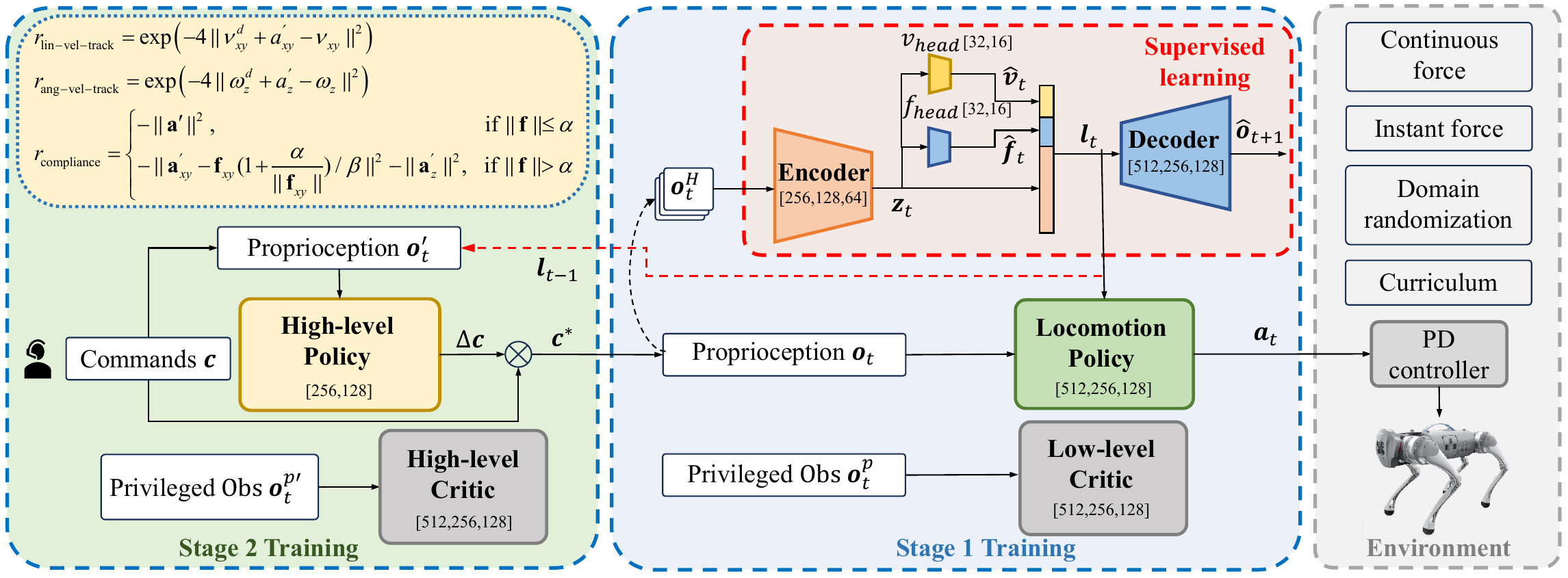}
\caption{HAC-LOCO framework}
 \label{fig: framework}
\end{figure*}

\subsection{Learning-based compliance control}
Many works have focused on achieving compliant locomotion through learning-based methods. Much of the work identifies key factors influencing robot compliance, including reward design, controller parameters, and frequency. Most learning-based approaches integrate action rate penalties, torque penalties, and energy-efficient reward terms to penalize high torque output and encourage compliant locomotion in response to disturbances \cite{Lee2020Learning, Ji2022Concurrent, Xiao2024PALOCO}. Some studies emphasize that a low proportional gain is crucial to increasing robot compliance and thus reducing potential harm to the joint motors \cite{Kumar2021RMA, Hwangbo2019Learning}. Additionally, Gangapurwala \emph{et. al.} claims that learning a low-frequency locomotion controller can enhance robot robustness and compliance with external disturbances \cite{Gangapurwala2023Learning}. However, the aforementioned works primarily focus on passive compliance with disturbances.

Some studies specifically address robot compliance for balance recovery in response to external disturbances. Lee \emph{et. al.} introduced a framework for the simulation and control of humanoid robots that enables physically compliant interactions with the environment \cite{Lee2022Deep}. However, this architecture requires sensing interaction forces, which makes its real-world implementation complicated. Hartmann \emph{et. al.} proposed a multi-stage episodic reinforcement learning approach for training a compliant quadruped controller \cite{Hartmann2024Deep}. This approach incorporates explicit recovery stages, where tracking rewards are given irrespective of the agent’s motions. This allows the agent to recover compliantly from disturbances before resuming task-specific locomotion, rather than rigidly following predefined commands. However, this work does not account for robots subjected to continuous external forces. Locomotion compliance is significantly affected by a recovery time parameter, but it is challenging to determine an optimal value. Hence, it remains a challenge to make compliant response based on disturbance estimation.



\section{Methodology}
\label{sec: Methodology}
This section presents the HAC-LOCO framework, which consists of two key components (as shown in Fig. \ref{fig: framework}): (1) a low-level robust locomotion policy that uses a historical feature encoder to accurately estimate body velocity and force disturbances based on proprioceptive feedback, and (2) a high-level lightweight compliance policy that adjusts velocity commands based on proprioceptive observations and feedback from the low-level controller, enabling natural compliance to external disturbances. This hierarchical architecture decouples disturbance sensing from compliant behavior generation. It is trained in two phases, with the low-level policy trained first, followed by the high-level policy.

\subsection{Low-level locomotion policy}
\label{sec: Low-level motion controller}
\subsubsection{\textbf{Policy Observation}}
The low-level locomotion policy observes a historical sequence of proprioceptive information, denoted as $\mathbf{o}_t^H = [o_{t-H+1}, \dots, o_t]^T$, with $H = 10$ in this work. The observation at time step $t$ includes proprioceptive feedback, trunk velocity commands, and the previous policy action, represented as $\mathbf{o}_t = [\mathbf{s}_t, \mathbf{c}_t, a_{t-1}]^T$. Specifically, the proprioceptive feedback $\mathbf{s}_t$ consists of angular velocity $\omega_t$, the projected gravity vector $g_t$, joint positions and velocities  on this step and previous step $q_t, \dot{q}_t, q_{t-1}, \dot{q}_{t-1}$, and the clock signal $\mathbf{t}_t = [\sin(2\pi f t), \cos(2\pi f t)]$ with $f = 2.5$ (desired step frequency). The velocity command $\mathbf{c}_t$ is expressed as $\mathbf{c}_t = [v_{x}^{d}, v_{y}^{d}, \omega_{z}^{d}]^T$, where $v_{x}^{d}$ and $v_{y}^{d}$ denote the horizontal trunk linear velocities command, and $\omega_{z}^{d}$ is the trunk angular velocity command.

\subsubsection{\textbf{Network Architecture}}
The architecture of the low-level locomotion policy is illustrated in Fig. \ref{fig: framework}. The encoder module $\mathit{E}$ processes historical proprioceptive observations, encoding them into a low-dimensional feature vector $\mathbf{z}_t = \mathit{E}(\mathbf{o}_t^H)$. This vector is then passed through two distinct heads: the force-head $f_{\text{head}}$, which estimates the external force $\hat{\mathbf{f}}_t = f_{\text{head}}(\mathbf{z}_t)$, and the velocity-head $v_{\text{head}}$, which estimates the body velocity $\hat{\mathbf{v}}_t = v_{\text{head}}(\mathbf{z}_t)$. The outputs of both heads are concatenated to form the latent representation $\mathbf{l}_t = [\mathbf{z}_t, \hat{\mathbf{f}}_t, \hat{\mathbf{v}}_t]^T$. This latent feature is then passed through a decoder module to predict the next-time-step observation $\hat{\mathbf{o}}_{t+1} = \mathit{D}(\mathbf{l}_t)$. Finally, the concatenated feature $\mathbf{l}_t$ and the current observation $\mathbf{o}_t$ are input into the policy network $\pi$, which generates the action $\mathbf{a}_t = \pi(\mathbf{o}_t, \mathbf{l}_t)$. The output action $\mathbf{a}_t \in \mathbb{R}^{12}$ represents target joint positions, which are adjusted and scaled before being tracked by joint PD controllers with gains $K_p = 30$ and $K_d = 0.75$.

\subsubsection{\textbf{Training Process}}
\label{sec: Training Process Low Level}
The low-level locomotion policy is trained using a combination of supervised learning and reinforcement learning. We apply the Proximal Policy Optimization (PPO) algorithm  \cite{Schulman2017Proximal} for reinforcement learning. The loss function of PPO is integrated with that of the supervised learning component to update the policy network parameters. Supervised learning facilitates the encoder in extracting critical information about external disturbances. Specifically, the velocity-head and force-head are trained to estimate the trunk velocity and external forces, respectively. A decoder is trained simultaneously to reconstruct the robot's dynamics from the latent representation by the encoder, which encourages the encoder to capture pivotal features from raw historical data. The supervised learning loss consists of two estimation losses and one reconstruction loss.
\begin{equation} 
\mathit{L}_{\text{sup}} = \omega_1\|\mathbf{v}_t - \hat{\mathbf{v}}_t\|^2 + 
\omega_2\|\mathbf{f}_t - \hat{\mathbf{f}}_t\|^2 
+ \omega_3\|\mathbf{o}_{t+1} - \hat{\mathbf{o}}_{t+1}\|^2, 
\end{equation}
where $\omega_1$, $\omega_2$, and $\omega_3$ are weights to balance different loss elements. The total loss for policy training is the sum of the supervised learning loss and the PPO loss, so
\begin{equation}
    \mathit{L} = \mathit{L}_{\text{sup}} + \mathit{L}_{\text{PPO}}.
\end{equation}

The RL reward function is a weighted sum of task and auxiliary rewards, as shown in Table \ref{table: RL_Reward}. In this table, $E$ and $V$ represent the mean and variance, respectively, and $\beta_i$ is the duty factor for leg $i$. The power distribution term, $V(\tau\dot{q})_h + V(\tau\dot{q})_t + V(\tau\dot{q})_c$, refers to the sum of the joint power variance across the hip, thigh, and calf joints.

An asymmetric actor-critic architecture  \cite{Pinto2018Asymmetric} is used, where the critic network receives a more detailed observation. The critic’s input is $\mathbf{o}_t^p = [\mathbf{o}_t, \mathbf{v}_t, \mathbf{d}_t, \mathbf{f}_t^H, \mathbf{h}_t]^T$, with $\mathbf{v}_t$ representing the trunk velocity, $\mathbf{d}_t$ containing environment properties (randomized via domain randomization as described in Section \ref{sec: implimentation details}, including ground friction, trunk mass, and motor gain), $\mathbf{f}_t^H = [f_{t-H+1}, \dots, f_t]^T$ representing historical external forces in the trunk frame, and $\mathbf{h}_t$ being the terrain heightmap surrounding the robot.
\begin{table}[!t]
\centering
 \caption{Reward functions and weights}
 \label{table: RL_Reward}
\begin{tabular}{lcc}
\toprule
Name&Expression&Weight\\
\hline
Linear velocity tracking&$\exp(-\frac{||v_{b,xy}^* - v_{b,xy}||^2}{0.25})$ &1.0\\
Angular velocity tracking&$\exp(-\frac{||\omega_{b,z}^* - \omega_{b,z}||^2}{0.25})$&0.5\\
Linear velocity penalty&$v_{b,z}^2$ &-2.0\\
Angular velocity penalty&$||\omega_{b,xy}||^2$&-0.05\\
Trunk orientation&$||\hat{g}_{x}||^2+||\hat{g}_{y}||^2$&-1.0\\
Trunk height&$\exp(-\frac{||h_b^* - h_b||^2}{0.01})$&0.5\\
Joints acceleration&$||\frac{\dot{q}_{j-1}-\dot{q}_{j}}{dt}||^2$&-2.5e-7\\
Joints torque&$||\tau_{j}||^2$&-1e-4\\
Power distribution&$V(\tau\dot{q})_{h}+V(\tau\dot{q})_{t}+V(\tau\dot{q})_{c}$&-1e-10\\
Action rate-1&$||q^*_{j-1}-q^*_{j}||^2$&-0.01\\
Action rate-2&$||q^*_{j-2}-2q^*_{j-1}+q^*_{j}||^2$&-0.01\\
Self collision&$n_{collisions}$&-0.2\\
Foot air time&$\sum^4_{f=0}min(t_{air,f}-0.4,0)$&1.0\\
Foot swing height&$\sum^4_{n=1}\exp(-\frac{||p_{f}-p_{d}||^2}{0.02})$&0.3\\
Stable stride&$V(\beta_{[0:3]})/E(\beta_{[0:3]})$&1.0\\
\bottomrule
\end{tabular}
\end{table}

\subsection{High-level compliance policy}
\label{sec: High-level compliance policy}

\subsubsection{\textbf{Policy Observation and Action}}
The high-level compliance policy receives latent information from the encoder of the low-level locomotion policy. In return, it generates a residual velocity command to facilitate compliant behavior in response to external force disturbances. This design follows the philosophy of conventional impedance control that adjusts a reference velocity based on external force feedback. To avoid confusion with the low-level policy, we denote the high-level policy's action as $\mathbf{a}_t'$ and its observation as $\mathbf{o}_t'$. The observation space for the high-level policy includes the robot's proprioceptive feedback $\mathbf{s}_t$, the original trunk velocity command $\mathbf{c}_t$, the last latent feature $\mathbf{l}_{t-1}$, and the previous actions from both the high-level and low-level policies, $\mathbf{a}_{t-1}'$ and $\mathbf{a}_{t-1}$, respectively.  Hence, one has
\begin{equation}
    \mathbf{o}_t' = [\mathbf{s}_t, \mathbf{c}_t, \mathbf{l}_{t-1}, \mathbf{a}_{t-1}', \mathbf{a}_{t-1}].
\end{equation}

The action $\mathbf{a}_t'$ represents the residual velocity commands to facilitate compliant behavior, given by $\mathbf{a}_t' = [\Delta v_x^d, \Delta v_y^d, \Delta \omega_z^d]^T$. The final command to the low-level policy is the sum of the original velocity command and the residual velocity command, expressed as
\begin{equation}
    \mathbf{c}_t^* = \mathbf{c}_t + \mathbf{a}_t',
\end{equation}
where $\mathbf{c}_t^*$ is the modified velocity command to be tracked by the low-level policy.

\subsubsection{\textbf{Training Process}}
\label{sec: Training Policy - high level}

The high-level policy is trained in the second stage after the low-level policy is complete. During this phase, the low-level policy is frozen and treated as a part of the environment. The PPO algorithm with an asymmetric actor-critic architecture is applied to train this module. The privileged observation for the critic network, $\mathbf{o}_t^{p'}$, is given by $\mathbf{o}_t^{p'} = [\mathbf{o}_t', \mathbf{v}_t, \mathbf{d}_t, \mathbf{f}^H_t, \mathbf{h}_t]$, where $\mathbf{v}_t$, $\mathbf{d}_t$, $\mathbf{f}_t^H$, and $\mathbf{h}_t$ are defined as in the low-level policy, as described in Section \ref{sec: Training Process Low Level}.


We make two modifications to the reward function. The first is to revise the velocity tracking rewards in Table \ref{table: RL_Reward}, while the second is to introduce a new resistance-compliance reward.
These adjustments to the overall reward function enable a quadruped robot to effectively reject minor force disturbances while exhibiting compliant behavior in response to larger ones. A compliance behavior is represented by a force threshold $\alpha$ and a virtual impedance $\beta$, where $\alpha$ denotes the force threshold that determines the boundary between precise tracking and compliant behavior. When the external force $|\mathbf{f}_t|$ exceeds this threshold, a robot transitions from precise tracking to compliant behavior. The parameter $\beta$ adjusts compliance flexibility with smaller values increasing compliance and larger values making the robot stiffer. The new velocity tracking rewards are thus adjusted to be
\begin{align}
    r_{\text{lin\_vel}} &= \exp\left(-4\|v^d_{xy} + a'_{xy} - v_{xy}\|^2\right),\\
    r_{\text{ang\_vel}} &= \exp\left(-4\|\omega^d_z + a'_z - \omega_z\|^2\right),
\end{align}
It should be noted that the low-level locomotion policy only needs to track the modified velocity commands. The newly introduced resistance-compliance reward is
\begin{equation*}
    r_{\text{comp}} =\left\{
    \begin{array}{ll}
    -\|\mathbf{a}'\|^2, & \text{if } \|\mathbf{f}\| \leq \alpha\\
    -\|\mathbf{a}'_{xy} - \frac{\mathbf{f}_{xy}}{\beta}(1+\frac{\alpha}{|\mathbf{f}_{xy}|})\|^2 -\|\mathbf{a}_{z}'\|^2, & \text{if } \|\mathbf{f}\| > \alpha
    \end{array}
    \right.
\end{equation*}
This formulation encourages the high-level policy to minimize command adjustments when external forces are below the threshold $\alpha$, and to generate compliant behavior with desired impedance $\beta$ (i.e., encouraging $\mathbf{a}'_{xy} \approx \frac{\mathbf{f}_{xy}}{\beta}$) when forces exceed $\alpha$. It combines the interpretability of model-based control (through the physical meaning of $\alpha$ and $\beta$) with the flexibility of reinforcement learning, optimizing stability and energy efficiency while ensuring a smooth transition from precise tracking to compliance.

The learned compliant behavior can be adjusted by changing $\alpha$ and $\beta$ in the reward function without necessitating a retraining of the low-level locomotion policy. This feature facilitates the efficient customization of locomotion controllers with varying characteristics. Additionally, the magnitude of $\mathbf{a}'_z$ is penalized within the resistance-compliance reward to uphold a robot's heading direction in the presence of external disturbances. This reward function can be extended to align a robot's heading with the direction of an external force, catering to specific applications like human-robot interaction. These varied compliant behaviors are showcased in Section \ref{sec: Results} through experimental validation, underscoring the versatility and effectiveness of the proposed approach.

\begin{table}[!b]
\centering
\caption{Domain Randomization Parameter}
\label{table: Domain Randomization}
\begin{tabular}{clc}
\toprule
&Parameter&Range\\
\hline
\multirow{4}{*}{\makecell{Randomized\\dynamics}}
&Trunk mass&[-1,3] kg\\
 & Trunk  inertia scale&[0.95, 1.05] \\
&COM displacement & [-0.03, 0.03] m\\
&Ground friction coefficient&[0.2, 1.2]\\
 & Ground restitution coefficient&[0, 0.2]\\
 \multirow{4}{*}{\makecell{Randomized\\Actuator}}& Joint $K_p$  factor&[0.8, 1.2]\\
 & Joint $K_d$  factor&[0.8, 1.2]\\1
 & Motor strength&[0.9, 1.1]\\
 & Action delay&[0, 20] ms\\

\multirow{4}{*}{\makecell{Sensor\\noises}}
&Trunk angular velocity noise&[-0.05, 0.05] rad/s\\
&Gravity vector noise&[-0.05, 0.05]\\
&Joint positions noise&[-0.01, 0.01] rad\\
&Joint velocities noise&[-0.075, 0.075] rad/s\\
\bottomrule
\end{tabular}
\end{table}

\subsection{Implementation details}
\label{sec: implimentation details}
The high-level policy shares the same implementation setup as the low-level one, including curriculum learning and domain randomization. For curriculum learning, a velocity curriculum and grid-based adaptive terrain curriculum, as proposed in  \cite{Margolis2024Rapid}, are employed to allow the policy to gradually adapt to increasingly complex commands and environments. The maximum velocity command ranges are set as \( v_x^* \in [-2, 2] \, \text{m/s} \), \( v_y^* \in [-1, 1] \, \text{m/s} \), and \( w_z^* \in [-2, 2] \, \text{rad/s} \)
The terrain types include smooth slopes, rough slopes, stairs (up and down), and discrete terrain features.

A domain randomization strategy is adopted to mitigate the simulation-to-reality gap, as outlined in Table \ref{table: Domain Randomization}. At training, we consider both impact force disturbances and persistent force disturbances. The impact force disturbances are simulated by applying random velocity and angular velocity impulses, $\Delta \mathbf{v}, \Delta \mathbf{\omega}$ to a robot every 10 seconds. Persistent external forces $ \mathbf{f} $ are continuously applied to the trunk of a robot, with their magnitude and direction being randomized every 4 seconds. 

For the low-level policy, curriculum-based disturbances are applied, with the maximum disturbance magnitude determined by the curriculum. In contrast, for training the high-level policy, disturbance magnitudes are always set to the upper limits defined by the curriculum. The maximum disturbance magnitudes are set as $ \|\mathbf{f}\|_{\text{max}} = 50 \, \text{N} $, $ \|\Delta \mathbf{v}\|_{\text{max}} = 1.5 \, \text{m/s} $, and $ \|\Delta \omega\|_{\text{max}} = 1.5 \, \text{rad/s} $.



\section{Experimental Results} \label{sec: Results} 
The training of both high-level and low-level policies is conducted in the Isaac Gym environment with 4096 parallel agents \cite{Makoviychuk2021Isaac}, using an episode length of 20 seconds with early termination triggered when the robot's trunk contacts the ground. The Proximal Policy Optimization (PPO) algorithm follows the hyperparameter setup detailed in \cite{Zhang2024SYNLOCOVE}, and the network architecture is depicted in Fig. \ref{fig: framework}. Experimental validation is performed on the Unitree Go1 quadruped robot, with both policies operating at 50 Hz during execution. In the first training stage, approximately 200 million samples are required, with convergence achieved in around 12 hours on a laptop equipped with an RTX 4060 GPU. The second stage requires only about 2 million samples for full convergence, benefiting from the lightweight nature of the compliance policy and the reusability of the low-level locomotion policy.

We implement the following algorithms for method comparison, ablation studies, and performance evaluation:

\begin{itemize}
    \item \textbf{Deep Compliant Control (DCC)} \cite{Hartmann2024Deep}: A learning-based compliance control with an additional recovery training.
    \item \textbf{PA-LOCO} \cite{Xiao2024PALOCO}: A perturbation-adaptive locomotion controller designed for robust locomotion.
    \item \textbf{HAC}: The purposed HAC-LOCO framework with both high-level and low-level policy.
    \item \textbf{HAC-Low}: Low level robust policy in the purposed HAC-LOCO framework.
\end{itemize}

\subsection{Compliant behavior toward continuous external forces}
\begin{figure}
    \centering
    \includegraphics[width=1\linewidth]{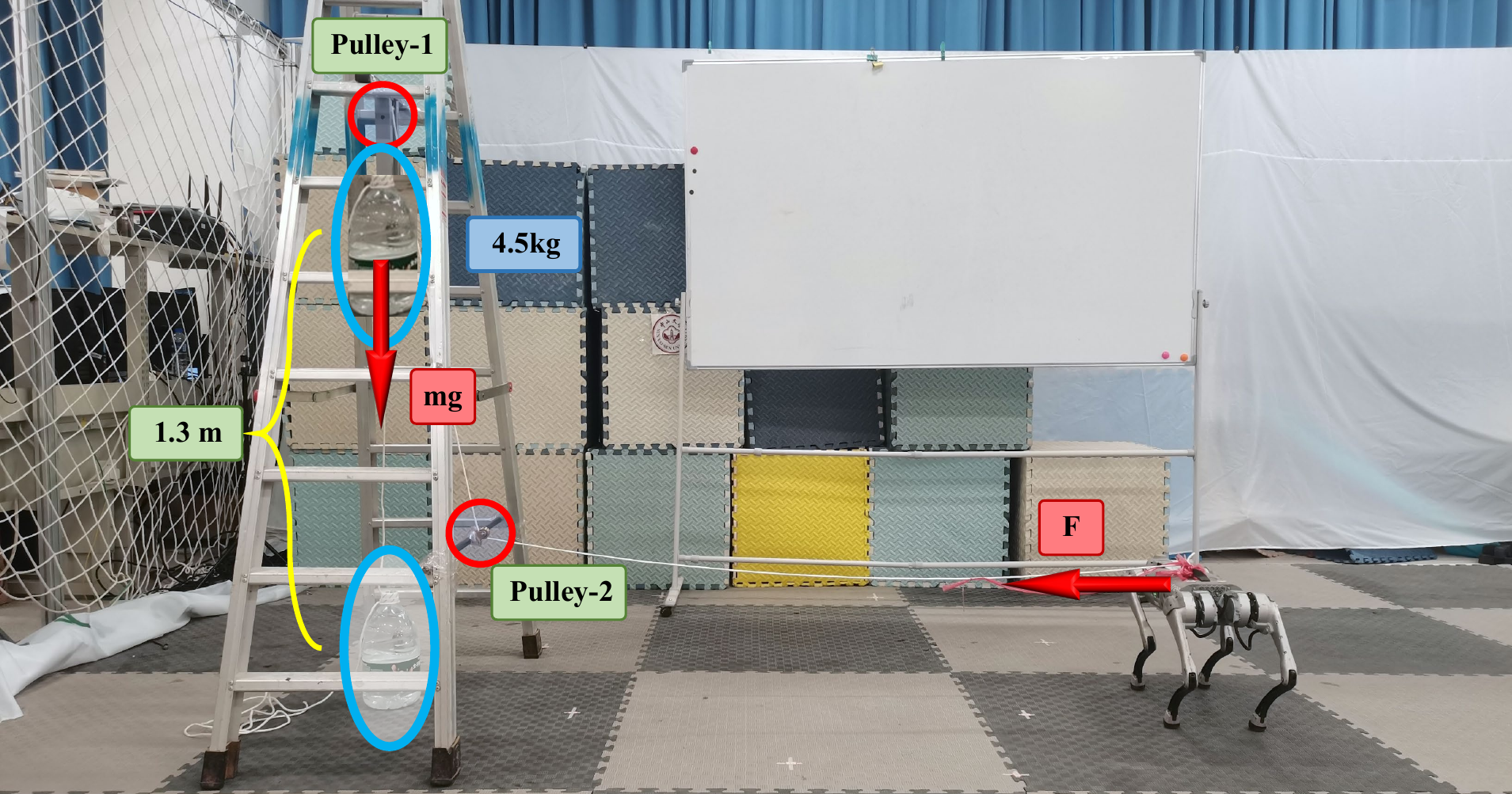}
    \caption{A cable-pulley system to generate force disturbances}
    \label{fig: cable-pulley system}
\end{figure}

A cable-pulley system, as illustrated in Fig. \ref{fig: cable-pulley system}, is designed to evaluate the effectiveness of the proposed HAC-LOCO framework in achieving compliant behavior under continuous external disturbances. This system can generate precise, continuous forces, ensuring reproducibility and consistency in applying force disturbances across trials. The setup consists of two fixed pulleys, a kernmantle rope, and $4.5$ L bottled water. The rope transmits the gravitational force exerted by the bottled water to the quadruped robot. During the experiment, the water tank is elevated and subsequently released, applying a sustained lateral force to the robot's trunk. 



\begin{figure}[t]
 \centering
 \includegraphics[width = \linewidth]{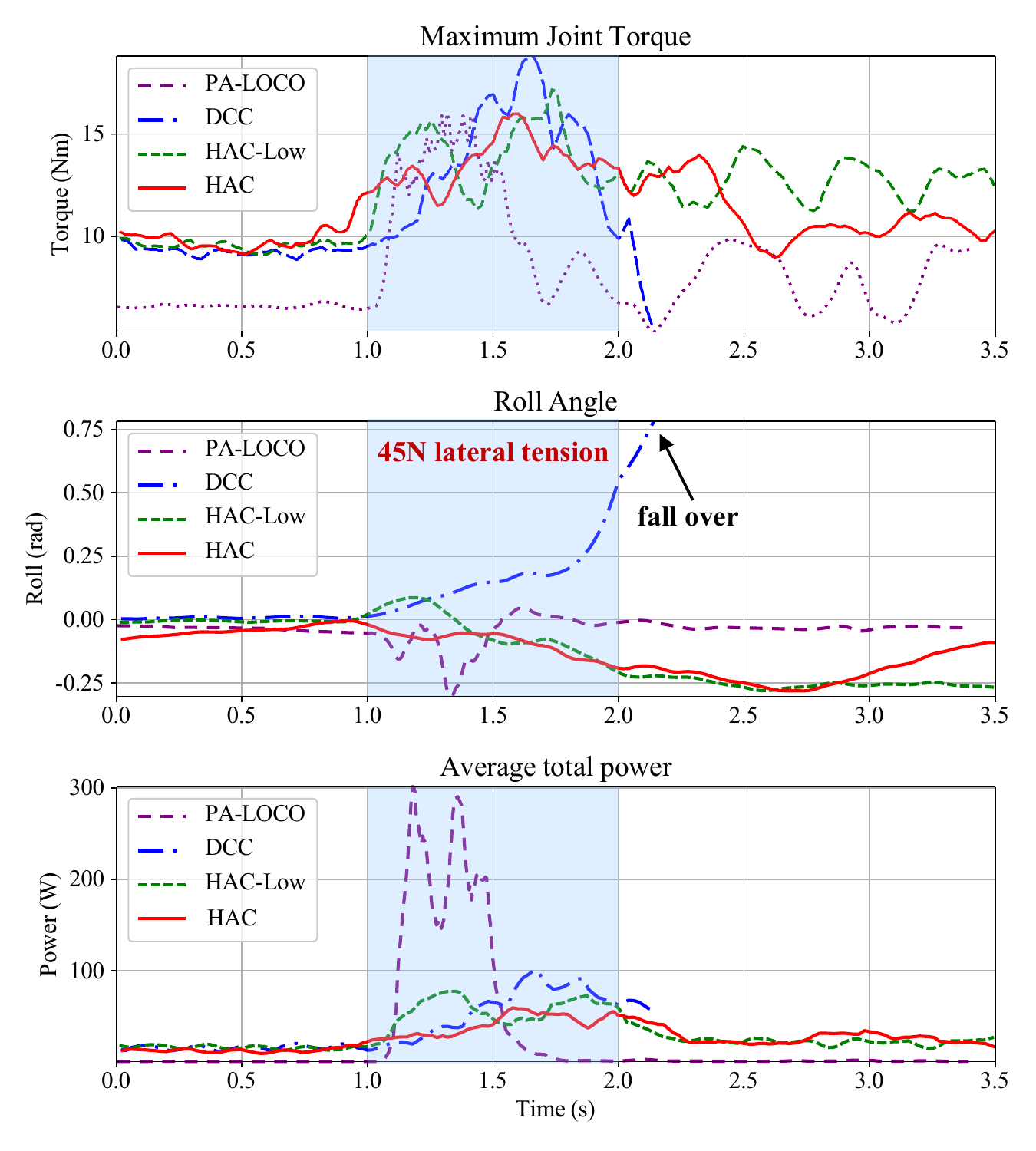}
 \caption{Compliant performance of different controllers at impact and persistent lateral forces. 
 }
 \label{fig: exp-1}
\end{figure}

Experiments were conducted in a scenario with a quadruped robot at a stationary condition when it is subject to a lateral impact followed by continuous external forces applied using a cable-pulley system. The compliance behavior of each controller is assessed based on three key performance metrics: maximum joint torque across all joints, trunk roll angle, and robot power consumption. The results, presented in Figure \ref{fig: exp-1}, show the performance of each controller under sustained disturbances in terms of stability, safety, and energy efficiency. During the experiment, the DCC controller can withstand the initial impact but fails to maintain stability under continuous disturbances, ultimately leading to a failure to remain upright. This instability is attributed to the control design neglecting sustained disturbances, thereby hindering the robot's ability to adapt its posture for balance maintenance. PA-LOCO shows resilience against instant and continuous external forces, displaying dynamic and aggressive locomotion with rapid leg swings for stabilization. This leads to sharp peaks in joint torque and notably higher power consumption post-impact.

In contrast, the low-level policy in the HAC-LOCO framework effectively mitigates external impacts, resulting in lower maximum joint torques and reduced total power consumption. However, it tends to remain rigid when facing continuous external forces, maintaining a fixed position without postural adaptation. With the aid of the residual velocity commands by a high-level policy based on estimated external forces, HAC-LOCO enhances compliance by dynamically adjusting its position in response to sustained disturbances instead of rigidly resisting them. This approach improves stability in trunk posture with smoother power consumption. Hence, HAC-LOCO shows the highest level of compliance among tested controllers, striking a balance between velocity tracking and compliance optimization.


Comprehensive simulation evaluations are performed to further assess policy performance under instantaneous and sustained disturbances in varied scenarios.  Each policy is tested over a $20$-second episode in the Isaac Gym environment with 4,096 parallel simulations on different setups. Random impulsive forces are applied every 3 seconds for impact force assessment, while sustained external forces were randomly reapplied at the same intervals for continuous force evaluation. The magnitudes of external forces and domain randomization settings are kept consistent with the training setup. The findings from the simulations are outlined in Table \ref{table: Metrics}.

The simulation results reveal that the HAC-Low policy attains the highest survival rate, demonstrating sufficient robustness against external disturbances. Its performance under instantaneous forces closely matches that of HAC, the full HAC-LOCO framework, suggesting that the low-level policy primarily governs responses to sudden impacts. However, in the face of continuous disturbances, the complete HAC-LOCO framework displays notably lower maximum torque compared to the low-level policy in isolation. This decrease affirms that the high-level policy effectively bolsters compliance when confronted with ongoing external forces. These outcomes align with those observed in physical trials, further affirming the efficacy of HAC-LOCO in enhancing compliant locomotion.

\begin{table}[htbp]
\centering
\caption{Simulation Results of Different Controllers}
\label{table: Metrics}
\begin{tabular}{lcccc}
\toprule
\multicolumn{5}{l}{\textbf{Instant Force}} \\
\textbf{Metrics} & \textbf{PA-LOCO} & \textbf{DCC} & \textbf{HAC-Low} & \textbf{HAC} \\
\midrule
Survival rate (\%) & 86 & 79 & 89 & 87 \\
Mean joint torque (Nm) & 44.6 & 29.1 & 32.8 & 31.0 \\
Mean robot power (W) & 156.3 & 72.6 & 75.9 & 81.2 \\
\midrule
\multicolumn{5}{l}{\textbf{Constant Force}} \\
\textbf{Metrics} & \textbf{PA-LOCO} & \textbf{DCC} & \textbf{HAC-Low} & \textbf{HAC} \\
\midrule
Survival rate (\%) & 74 & 62 & 85 & 83 \\
Mean joint torque (Nm) & 49.7 & 31.6 & 42.0 & 29.1 \\
Mean robot power (W) & 203.3 & 160.7 & 142.5 & 149.2 \\
\bottomrule
\end{tabular}
\end{table}


\subsection{Effectiveness of Low Level Network Design}

\begin{figure}
    \centering
    \includegraphics[width=1\linewidth]{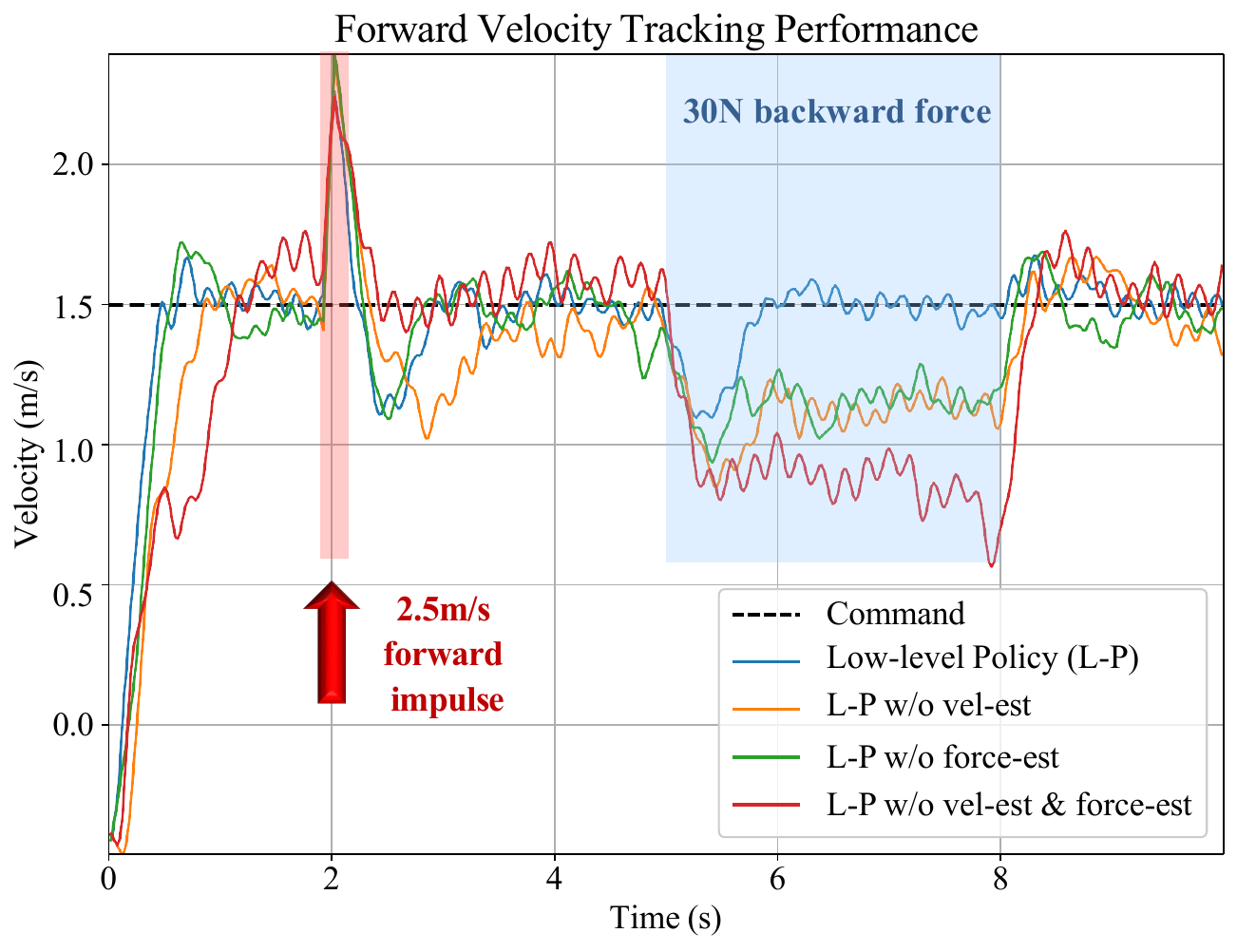}
    \caption{Comparison of velocity tracking performance with and without explicit estimation of external forces and velocity.}
    \label{fig:vel-tracking}
\end{figure}

A series of ablation studies are conducted in a simulated dynamic environment to validate the effectiveness of different modules of the low-level network design. Specifically, the full low-level motion policy (L-P) is compared against three modified versions: L-P without velocity estimation (L-P w/o vel-est), L-P without force estimation (L-P w/o force-est), and L-P without both force and velocity estimation (L-P w/o force-vel-est). These variants are trained by setting the corresponding supervised learning loss terms to zero while maintaining an identical number of training steps.


The evaluation is performed in a test environment where the robot tracks a forward velocity of $1.5$ m/s on rough terrain. At $t=2$ s, a forward impact force equivalent to an instant velocity change of $2.5$ m/s is applied, followed by a backward continuous force of $30$ N from $t=5$ s to $t=8$ s. The velocity tracking results are shown in Fig. \ref{fig:vel-tracking}. Experimental results demonstrate that the full low-level motion policy (L-P) achieves the highest velocity tracking accuracy, maintaining minimal estimation errors for both velocity and external force. When velocity estimation is removed (L-P w/o vel-est), tracking accuracy deteriorates after the impact. Similarly, removing force estimation (L-P w/o force-est) leads to reduced tracking accuracy under sustained forces. The most significant degradation occurs in the absence of both force and velocity estimation (L-P w/o vel-est \& force-est), where tracking performance degradation greatly under continuous disturbances. These findings suggest that explicit force and velocity estimation contribute to enhancing velocity tracking performance in environments with both instant and continuous disturbances.

\subsection{Effectiveness of High Level Network}
To assess the effectiveness of the high-level policy, we train two separate high-level policies with different compliance parameters: $\alpha = 10, \beta = 10$ (\emph{low impedance}) and $\alpha = 10, \beta = 30$ (\emph{high impedance}). These configurations allow us to analyze the influence of varied impedance on the robot’s compliance behavior, where lower $\beta$ correspond to higher compliance, making the robot more responsive to external forces, while higher $\beta$ values result in a stiffer response.

\begin{figure}[t]
    \centering
    \includegraphics[width=1\linewidth]{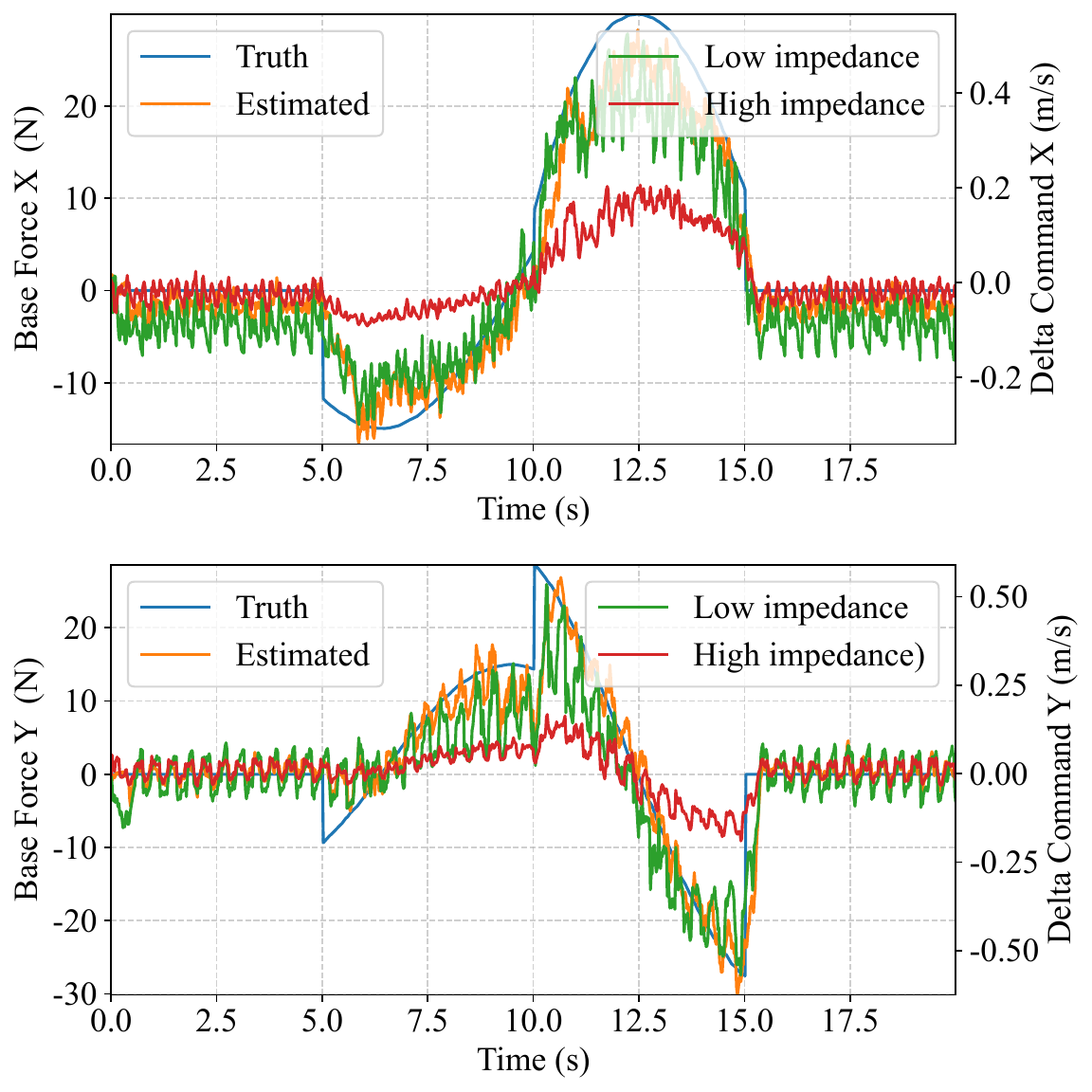}
    \caption{High-level policy output in response to external forces applied to the robot's trunk. The correction output is approximately proportional to the external force, demonstrating impedance-like behavior. By tuning the compliance parameters $\alpha$ and $\beta$ in the reward function during training, the virtual impedance characteristics of the policy can be modulated, enabling adaptive compliance control.}
    \label{fig:Experiment-3}
\end{figure}

To assess the compliance characteristics, the robot is commanded to track a forward velocity of $1\text{m/s}$ and an angular velocity of $0.5 \text{rad/s}$ on a rough flat surface. At $t=5$ s, a sustained external force $\mathbf{f} = [15, 0, 0] \, \text{N}$ is applied in the world frame, and at $t=10$, the external force changed to  $[30, 0, 0] \, \text{N}$. The estimated external force and the output of the command correction module are shown in Fig. \ref{fig:Experiment-3}. To ensure consistent motion states across evaluations, the correction output is recorded but not fed into the low-level motion network.

The results indicate that although the external force is time-varying in the robot’s body frame due to motion dynamics, the high-level policy is able to estimate the applied force with reasonable accuracy, though a minor delay is observed at the instant of the force application. The correction output from the high-level policy aligns with the direction of the external force and is approximately proportional in magnitude, suggesting that the high-level network effectively runs as impedance control. Under identical motion and external force conditions, a higher value of $\beta$ leads to a smaller correction output, implying a lower compliance level. Hence, by tuning $\beta$, the robot can achieve different impedance characteristics, offering a clear physical interpretation of the compliance parameters. These findings confirm that the high-level command correction module effectively regulates compliance behavior, allowing tunable and interpretable impedance control for quadruped locomotion.

\subsection{Locomotion in Various Outdoor Environments with Different Compliance}

    \begin{figure}[!t]
     \centering
     \includegraphics[width = \linewidth]{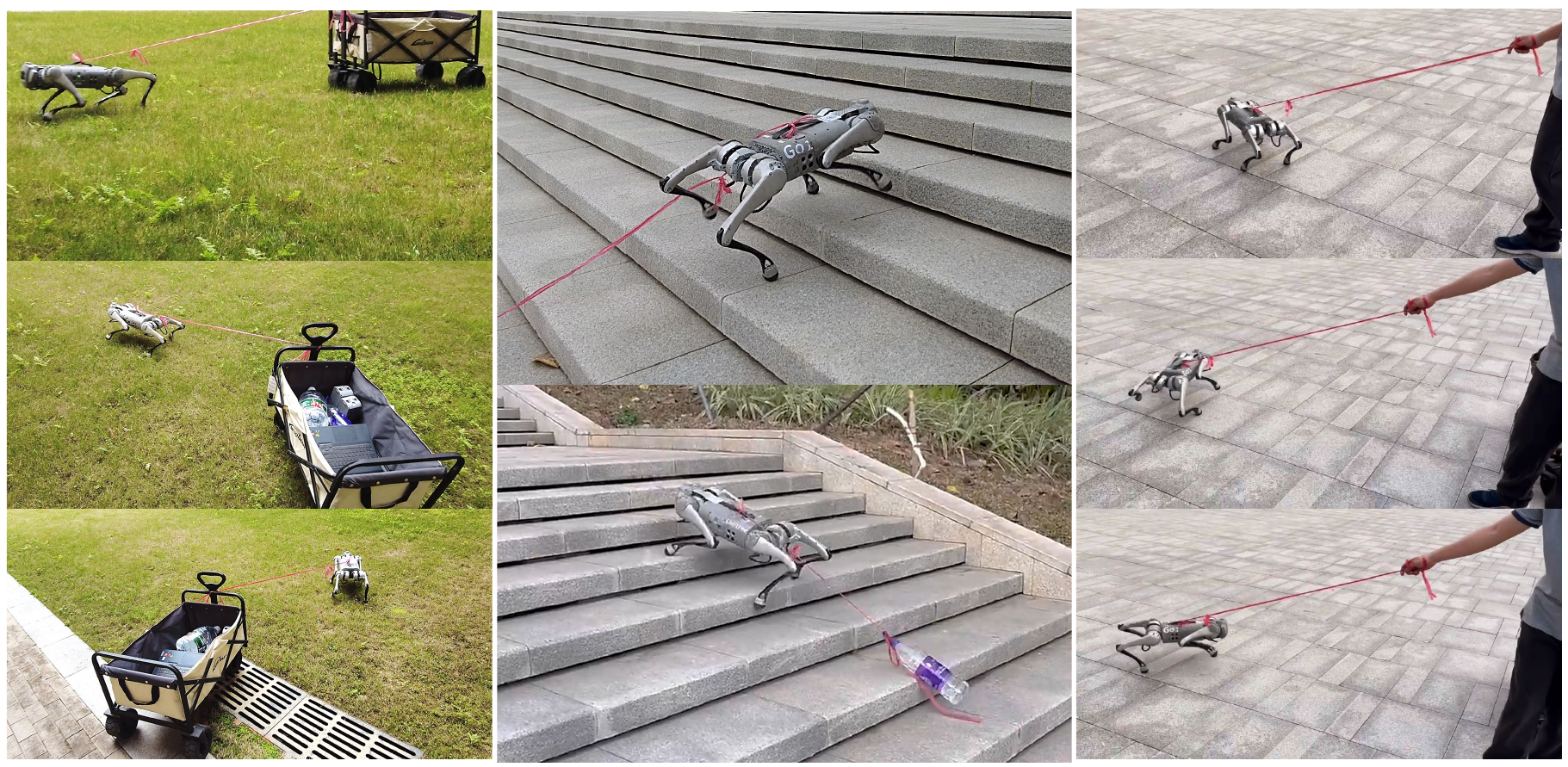}
     \caption{Outdoor experiments demonstrating the versatility of HAC-LOCO on different terrains.
     }
     \label{fig: outdoor experiments}
    \end{figure}


As shown in Fig. \ref{fig: outdoor experiments}, the proposed HAC-LOCO framework is validated in diverse outdoor environments, showcasing its adaptability to different compliance behaviors. The robot is initially tested on grassland, where controllers with higher impedance enable it to tow a picnic cart efficiently, while controllers with lower impedance allow a human operator to drag the quadruped and guide its heading direction with ease. The quadruped is further evaluated on stair terrain, where a high-impedance policy facilitates stable locomotion while dragging a large water bottle upstairs. Meanwhile, a lower-impedance policy enables the robot to follow human guidance when ascending and descending stairs. These scenarios validate the robustness of the HAC-LOCO low-level policy in handling various terrains and external disturbances, as well as the versatility of the high-level policy in achieving diverse compliant behaviors suitable for real-world applications.


\section{Conclusions} \label{sec: Conclusion}
This paper presented HAC-LOCO, a hierarchical reinforcement learning framework for active compliance control in quadruped locomotion. By integrating a robust low-level motion policy with a high-level compliance module, the framework enables the robot to resist impact forces while exhibiting compliant responses to sustained external disturbances. A learning-based estimator combining an AutoEncoder with explicit force and velocity estimation significantly improved disturbance estimation accuracy, enhancing velocity tracking and stability. Additionally, the lightweight compliance module, inspired by impedance control, allows for dynamic compliance adjustments without retraining the entire framework. Extensive simulations and physical experiments demonstrated that HAC-LOCO achieves superior compliance, reducing peak joint torques and power consumption while maintaining stability under external forces. The ability to modify compliance characteristics by adjusting parameters further highlights its adaptability. Future work will focus on adaptively selecting the force threshold and virtual impedance based on sensory feedback, improving the robot’s intelligence for actively responding to disturbances in unstructured environments.

\bibliography{ref}
\bibliographystyle{IEEEtran}

\end{document}